\documentclass[10pt,twocolumn,letterpaper]{article}

\usepackage[pagenumbers]{cvpr} 
\pdfoutput=1
\usepackage{graphicx}
\usepackage{amsmath}
\usepackage{amssymb}
\usepackage{booktabs}
\usepackage[table,xcdraw]{xcolor}
\usepackage{color}
\usepackage{stfloats}
\usepackage{float}
\usepackage{stfloats}
\usepackage{multirow} 
\usepackage{enumitem}
\usepackage[pagebackref,breaklinks,colorlinks]{hyperref}

\usepackage[capitalize]{cleveref}
\crefname{section}{Sec.}{Secs.}
\Crefname{section}{Section}{Sections}
\Crefname{table}{Table}{Tables}
\crefname{table}{Tab.}{Tabs.}

\def\cvprPaperID{7187} 
\def\confName{CVPR}
\def\confYear{2023}

\begin{document}

\title{Diverse 3D Hand Gesture Prediction from Body Dynamics by\\
 Bilateral Hand Disentanglement}

\vspace{-2mm}
\author{Xingqun Qi$^{1,4}$, Chen Liu$^{2}$, Muyi Sun$^{3}$, Lincheng Li$^{4}$, Changjie Fan$^{4}$, Xin Yu$^{1, 2}$\footnotemark[1]\\
$^{1}$ AAII, University of Technology Sydney  $^{2}$ The University of Queensland\\
$^{3}$ CRIPAC, NLPR, Institute of Automation, Chinese Academy of Sciences\\
$^{4}$ Netease Fuxi AI Lab\\
{\tt\small \{xingqunqi, yenanliu36\}@gmail.com, muyi.sun@cripac.ia.ac.cn  } \\
{\tt\small \{lilincheng, fanchangjie\}@corp.netease.com, xin.yu@uq.edu.au} 
}


\vspace{-2mm}
\twocolumn[{%
\renewcommand\twocolumn[1][]{#1}%
\maketitle
\vspace{-10mm}
\begin{center}
    \centering
    \includegraphics[width=0.85\linewidth]{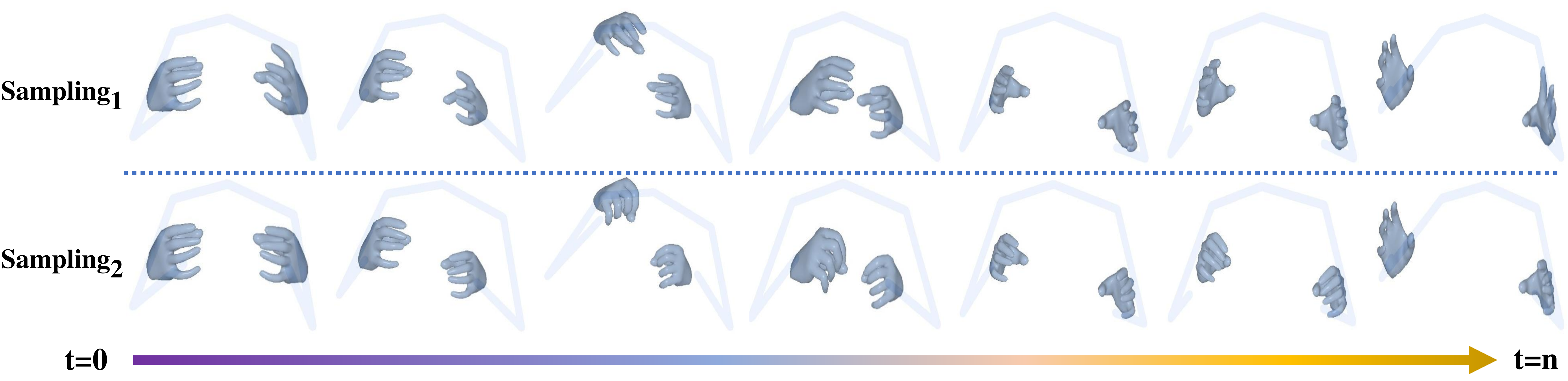}
    \vspace{-1em}
    \captionof{figure}{Diverse exemplary clips sampled by our method from \textbf{our newly collected TED Hands Dataset}. The vital frames are visualized to demonstrate that the hand gestures change with the upper body dynamics, synchronously. Best view on screen. }
    \label{fig:figure1}
\end{center}
}]
\renewcommand{\thefootnote}{\fnsymbol{footnote}}
\footnotetext[1]{Corresponding author.}

\begin{abstract}
\vspace{-1em}
Predicting natural and diverse 3D hand gestures from the upper body dynamics is a practical yet challenging task in virtual avatar creation. Previous works usually overlook the asymmetric motions between two hands and generate two hands in a holistic manner, leading to unnatural results. In this work, we introduce a novel bilateral hand disentanglement based two-stage 3D hand generation method to achieve natural and diverse 3D hand prediction from body dynamics. In the first stage, we intend to generate natural hand gestures by two hand-disentanglement branches. Considering the asymmetric gestures and motions of two hands, we introduce a Spatial-Residual Memory (SRM) module to model spatial interaction between the body and each hand by residual learning. To enhance the coordination of two hand motions wrt. body dynamics holistically, we then present a Temporal-Motion Memory (TMM) module. TMM can effectively model the temporal association between body dynamics and two hand motions. The second stage is built upon the insight that 3D hand predictions should be non-deterministic given the sequential body postures. Thus, we further diversify our 3D hand predictions based on the initial output from the stage one. Concretely, we propose a Prototypical-Memory Sampling Strategy (PSS) to generate the non-deterministic hand gestures by gradient-based Markov Chain Monte Carlo (MCMC) sampling. Extensive experiments demonstrate that our method outperforms the state-of-the-art models on the B2H dataset and our newly collected TED Hands dataset. The dataset and code are available at \href{https://github.com/XingqunQi-lab/Diverse-3D-Hand-Gesture-Prediction}{\textit{Diverse-3D-Hand-Gesture-Prediction}}.
\end{abstract}

\vspace{-5mm}
\section{Introduction}

Given a sequence of upper body skeletons, our task aims at predicting natural and diverse 3D hand gestures. Such non-verbal body-hand coordination plays an important role in various virtual avatar scenarios, including human-agent interface \cite{koppula2013anticipating,huang2022proxemics,wolfert2022review,salem2012generation, wang2022self}, co-speech gesture synthesis \cite{liu2022learning,liang2022seeg,yoon2019robots,yoon2020speech,fu2023eccvw}, holoportation \cite{orts2016holoportation}. 

However, it is quite difficult to predict the natural and diverse 3D hand gestures due to three major challenges: \textbf{(1) Spatially asymmetric motions:} Asymmetric motions of two hands have been overlooked by previous works \cite{Ng_2021_CVPR}, \emph{e.g.,} when one hand moves, the other could be static or moving slowly. 
\textbf{(2) Temporal consistency wrt. body dynamics:} Predicted sequential hand gestures should be temporally consistent with respect to the body dynamics. \textbf{(3) Non-deterministic hand prediction:} Given the upper body dynamics, various 3D hand gestures can match the body postures rather than a deterministic result. Since the existing dataset has only a few avatar identities, the generated hands often lack diversity.



Due to complex body-hand interaction and asymmetric movements of two hands, directly predicting the natural and diverse 3D hand is rather difficult. Therefore, we propose to address the aforementioned challenges in a prediction followed by a diversification paradigm. To be specific, we propose a novel bilateral hand disentanglement based two-stage method, to generate natural and diverse 3D hand gestures from upper body dynamics.

In the first stage, we aim to predict the natural 3D hand gestures. Our key insight is to establish the bilateral hand disentanglement for asymmetric motions of two hands. We construct two hand-disentangled branches that interact with a body-specific branch to initially predict natural 3D hands. 
Here, we leverage a single-hand autoencoder to extract each hand feature respectively, thus achieving the effects of two hands disentanglement. 
Furthermore, we propose a Spatial-Residual Memory (SRM) module to model the spatial relation between the body and each hand via residual learning. Specifically, we employ a spatial memory bank to store the spatial residual deformation representations of each hand. Then, we leverage the current upper body embedding as a \emph{query} to retrieve the most relevant spatial residual deformation. The queried spatial residual deformation and the synchronized hand embedding vector are used to generate the next step hand representation. 

To ensure the motions of both hands are temporally consistent with the upper body sequence, we present a Temporal-Motion Memory (TMM) module. TMM models the correlation between body dynamics and hand gestures. Similar to SRM, we utilize a temporal memory bank to store the motion features of each hand at the sequence level.
Moreover, we employ a motion encoder to acquire the input body dynamics representation, and then use the body dynamics representation as a \emph{query} to produce temporally consistent hand motion features.
In this fashion, we can effectively synchronize hand motions with body dynamics.


In the second stage, we develop a Prototypical-Memory Sampling Strategy (PSS) to diversify our 3D hand predictions based on the initial prediction at stage one. 
Here, we leverage an external memory that stores realistic hand prototypes, and then search a prototype that is closest to our initial predicted hands. Once we obtain the prototype, we project it into the feature space. To increase the diversity while preserving the authenticity of generated hand gestures, we perturb the prototype feature via gradient-based Markov Chain Monte Carlo (MCMC) sampling \cite{liu2001monte}. Specifically, a random noise is first sampled from a Gaussian distribution as a prior perturbation and then we update its posterior via Langevin dynamics based MCMC \cite{neal2011mcmc, dubey2016variance}. Then, the perturbation is concatenated with the prototype feature to produce realistic new hand gestures. 


Moreover, the existing 3D hand prediction dataset \cite{Ng_2021_CVPR} contains less than 10 avatar identities, resulting in insufficient diversity of gestures. Therefore, we newly collect a large-scale 3D hand gestures dataset (dubbed TED Hands) with more than 1.7K avatar identities from in-the-wild scenarios. Our dataset contains around 100 hours of TED talk speeches, enabling research on diverse 3D hand gesture predictions.
Extensive experiments conducted on the B2H dataset \cite{Ng_2021_CVPR} and our TED Hands dataset demonstrate our method outperforms the state-of-the-art.

\noindent Overall, our contributions are summarized as follows:
\begin{itemize}[leftmargin=*]
\vspace{-0.6em}
    \item We propose a novel bilateral hand disentanglement based two-stage 3D hand generation method to predict diverse 3D hand gestures from body dynamics in a prediction followed by diversification paradigm.
    \vspace{-0.6em}
    \item We present a Spatial-Residual Memory (SRM) module to predict authentic hand poses from disentangled hand representations and a Temporal-Motion Memory (TMM) module to ensure temporal consistency of generated 3D hands.
    \vspace{-0.6em}
    \item We design a Prototypical-Memory Sampling Strategy (PSS) to diversify our initial 3D hand prediction, thus obtaining natural and diverse hand gestures.
    \vspace{-0.6em}
    \item We collect a new large-scale sequential 3D hand dataset from 1.7K persons' hand gestures, significantly facilitating research on diverse 3D hands generation.
\end{itemize}

\section{Related Work}
Since our method predicts the 3D hand motions from upper body skeletons, we briefly review the approaches of 3D hand pose estimation and 3D human motion prediction.

\vspace{0.5em}
\noindent{\textbf{3D hand pose estimation:}}
3D hand estimation has received impressive attention for its wide applications in virtual reality, augmented reality, and robotics. Recently, parametric 3D hand model MANO \cite{romero2017embodied} and various mocap-captured 3D hand datasets \cite{chao2021dexycb,moon2020interhand2,hampali2020honnotate,
zimmermann2019freihand} signicantly improve the performance of 3D hand pose estimation from the single image \cite{li2022interacting, ge20193d, zhang2021interacting, spurr2021self, Park_2022_CVPR, zhou2020monocular}.
Among various image-based 3D hand estimation approaches, transformer-based models \cite{li2022interacting,lin2021end, Hampali_2022_CVPR} have demonstrated outstanding performance. 
For instance, Li \etal \cite{li2022interacting} propose an interacting attention graph transformer to estimate two-hand poses from a single image. 
Shreyas \etal \cite{Hampali_2022_CVPR} present a hand keypoint based transformer combined with convolutional networks to establish the association among different joints.

Since our task aims to predict the 3D hand motions from upper body skeletons, these single-image-based methods \cite{li2022interacting, lin2021end, Hampali_2022_CVPR} cannot be directly employed. However, their excellent performance in 3D hand estimation inspires us to model the body-hand interaction via transformer-based architectures. Besides, Ng \etal \cite{Ng_2021_CVPR} generate the hand gestures from body dynamics in a holistic manner. As the asymmetric motions of two hands, their method might produce some unnatural hand gestures. This also motivates us to disentangle two hands in hand pose estimation and then synchronize two-hands motions.


\begin{figure*}[t]
\begin{center}
\includegraphics[width=0.95\linewidth]{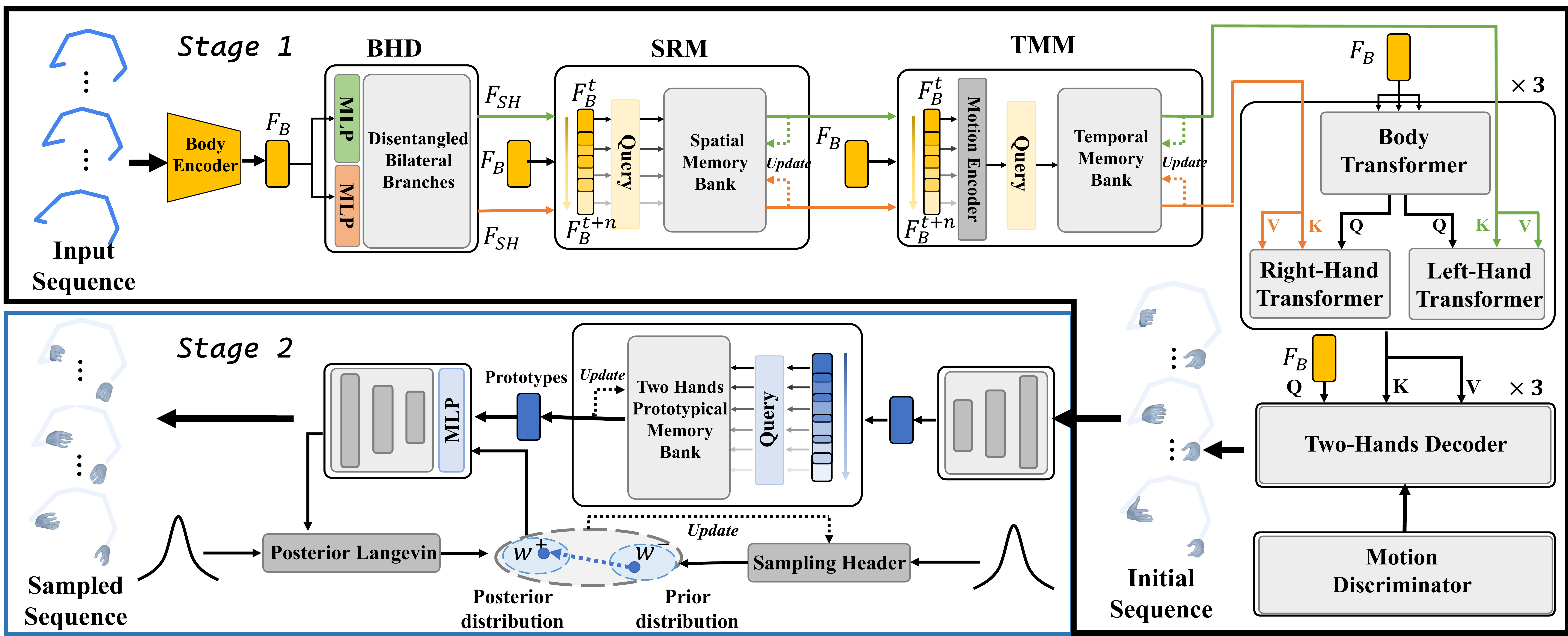}
\end{center}
\vspace{-1.5em}
\caption{The pipeline of our proposed method. In stage one, we generate the natural initial 3D hands from sequential upper body skeletons. In stage two, we diversify our 3D hand predictions based on the initial gestures at stage one.  }
\label{fig:figure2}
\vspace{-0.5em}
\end{figure*}

\vspace{0.5em}
\noindent{\textbf{3D human motion prediction:}}
Predicting human future motions from existing 3D poses has been widely studied \cite{wang2021multi,guo_2023_wacv, mao2019learning, 
ijcai2022p111, li2022skeleton}. Wang \etal \cite{wang2021multi} leverage a multi-range transformer that attends learn local-global features for multi-person motion prediction. 
Bouazizi \etal \cite{ijcai2022p111} introduce a multi-layer perception based human motion prediction network to learn spatio-temporal pose relation. 
Due to the similar nature between 3D human motion prediction and hand pose prediction, these models \cite{wang2021multi, guo_2023_wacv, mao2019learning, ijcai2022p111,li2022skeleton} can be easily adopted as our baselines. 
Moreover, we notice that memory based networks \cite{wu2022memvit, cheng2022xmem, yu2022memory, Wu_2022_BMVC} achieve promising performance in long-sequence modeling tasks. Motivated by this, we introduce external memories into our network to facilitate our long-range hand pose generation.


\section{Proposed Method}
\subsection{Problem Formulation}\label{Problem_Formulation}
Given a sequence of 3D upper body skeletons $\mathcal{B}=\left \{b^{t}  \right \} _{t=0}^{T} $, our goal is to predict the natural and diverse 3D hand poses $\mathcal{H}=\left \{h^{t}  \right \} _{t=0}^{T} $, where $T$ is the sequence length. Specifically, we define each step body skeleton (\emph{i.e.}, collars, shoulders, elbows, and wrists) $b^{t} \in \mathbb{R}^{24}$ as 8 joints. Each joint is represented by a 3D axis-angle representation. Meanwhile, the 3D hand pose at time $t$, $h^{t} \in \mathbb{R} ^{90}$, is described as 30 joints with 3D axis-angle representation. 

\subsection{Stage One: Natural Hand Prediction}\label{Natural_Hand_Prediction}
Due to complex body-hand interaction and asymmetric motions of two hands, 
we leverage the bilateral hand disentanglement branches and a body-specific branch to generate the natural 3D hands. Specifically, as depicted in Fig. \ref{fig:figure2} (stage one), our method mainly consists of a Bilateral Hand Disentanglement (BHD) module, a Spatial-Residual Memory (SRM) module, a Temporal-Motion Memory (TMM) module, and a transformer-based backbone. Firstly, we leverage a body encoder to obtain the body features $F_{\mathcal{B}}=\left \{ F_{\mathcal{B}}^{t}  \right \}_{t=0}^{T}$, where $F_{\mathcal{B}}^{t}  \in \mathbb{R} ^{1 \times C}$ denotes a body feature at time $t$, and $C$ is the feature dimension. Then, the body features are disentangled by BHD to acquire the single hand features $F_{\mathcal{SH}}=\left \{ F_{\mathcal{SH}}^{t}  \right \}_{t=0}^{T}$, where $F_{\mathcal{SH}}^{t}  \in \mathbb{R} ^{1 \times C}$.
Considering the asymmetry motions of two hands, we exploit SRM to model the spatial interaction between the body and each hand, respectively. Besides, we further develop TMM to keep the two hand gestures temporal consistent with body dynamics.
\begin{figure*}[t]
\begin{center}
\includegraphics[width=1\linewidth]{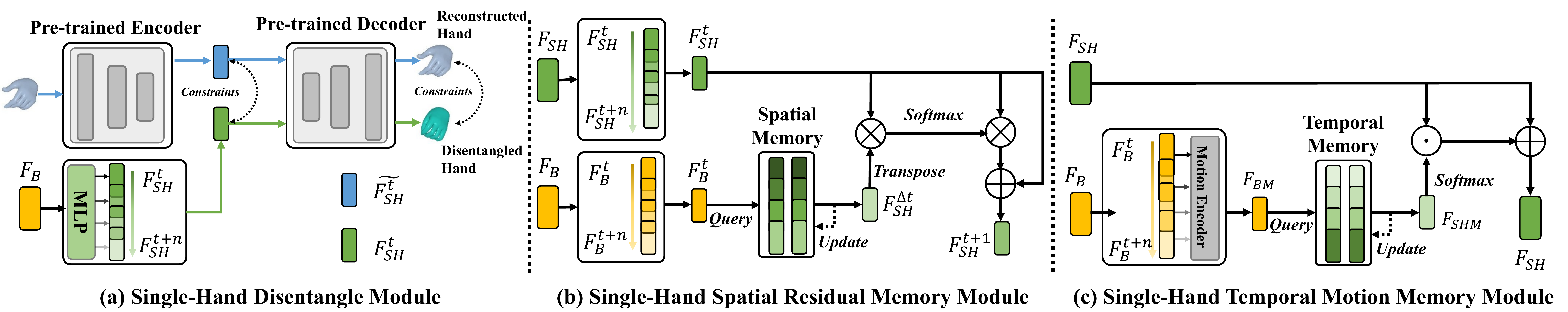}
\end{center}
\vspace{-1.5em}
\caption{Details of three sub-modules in the first stage. For brevity, we only show them in terms of single-branch view. }
\label{fig:figure3}
\end{figure*}

\vspace{0.5em}
\noindent{\textbf{Bilateral Hand Disentangle Module: }}
Due to the asymmetric motions of two hands, we employ a single hand autoencoder as the feature extractor to disentangle two hands. We firstly train an autoencoder on 3D single hands. As shown in Fig. \ref{fig:figure3} (a), we then utilize a Multilayer Perceptron (MLP) to project the body features into the single hand features $F_{\mathcal{SH}}$. To ensure these single hand features represent each disentangled hand separately, we conduct the feature level and 3D representation level constraints by the single hand autoencoder, as follows: 
\begin{equation}
\mathcal{L}_{dis} \!=\! \left \| \widetilde{F_{SH}^{t}} \!-\! F_{\mathcal{SH}}^{t} \right \|_{1}  \!+\! \left \| \mathbb{D}  \left( \widetilde{F_{SH}^{t}} \right) \!-\!\mathbb{D}  \left( F_{\mathcal{SH}}^{t} \right) \right \|_{1},  
\label{Eq1}
\end{equation}
where $\widetilde{F_{SH}^{t}}$ denotes reconstructed hand feature at time $t$,and $\mathbb{D}$ is the pre-trained decoder.

\vspace{0.5em}
\noindent{\textbf{Spatial-Residual Memory Module:}}
After obtaining the disentangled single hand features, we introduce a Spatial-Residual Memory (SRM) module  to model the spatial interaction between the body and each hand by residual learning. As illustrated in Fig. \ref{fig:figure3} (b), we leverage a spatial memory bank to store the spatial residual deformation representation between the current hand feature and the next step hand feature. Initially, each slot of the spatial memory bank is randomized as $ m_{j}\in \mathbb{R} ^{1\times C} $, where $j$ denotes slot index. To model the body-hand correlation, the current body feature is exploited as a \emph{query} to retrieve the most relevant memory slot by calculating the cosine similarity.

In particular, we represent the relevance between the query and memory slots via cosine similarity. Then we leverage the argmax operation to acquire the most relevant memory slot $m_{r}$. Intuitively, the retrieved most relevant memory slot is leveraged as the spatial residual deformation between the current feature and the next step hand feature. However, it is intractable for argmax operation to back-propagate gradients. Thus, we conduct the softmax operation on the above calculated cosine similarity matrix to obtain the slot-wise affinity matrix. Then we acquire the learnable spatial residual deformation $F_{\mathcal{SH}}^{\Delta t}$ by aggregation between the slot-wise affinity matrix and each memory slot. We apply the above reading process to allow gradient back-propagation. 
In the training phase, we utilize the current hand feature to \emph{update} the retrieved memory slot as $m_{r} \!=\! \gamma m_{r} + (1-\gamma) F_{\mathcal{SH}}^{t}$, where $\gamma \! = \! 0.8$ in our experiments. 

Different from obtaining the next hand feature by simple addition between the spatial residual deformation and current hand feature, we claim the spatial residual deformation refers to local spatial dependency from the current feature to the next time feature. This spatial dependency $S$ is formulated as: 
\begin{equation}
S_{t}^{t+1} \! = \! softmax({F_{\mathcal{SH}}^{\Delta t}}^{'} \otimes F_{\mathcal{SH}}^{t}),
\label{Eq2}
\end{equation}
where $\otimes$ indicates matrix multiplication, and $'$ indicates the transpose operation. With the help of the spatial dependency, the next step hand feature is represented as $F_{\mathcal{SH}}^{t+1} \!=\! F_{\mathcal{SH}}^{t} + F_{\mathcal{SH}}^{t} \otimes S_{t}^{t+1}$.
Through SRM, we obtain the spatial-enhanced single-hand feature.

\vspace{0.5em}
\noindent{\textbf{Temporal-Motion Memory Module:}}
We further propose a Temporal-Motion Memory (TMM) module to ensure the single hand motions consistent with the body dynamics, globally. As depicted in Fig. \ref{fig:figure3} (c), we develop a motion encoder to obtain the sequence-aware body motion embedding as $F_{\mathcal{BM}}\in\mathcal{R} ^{T\times 1} $. Here, we use body motion embedding to represent temporal changes among body skeletons in a sequence. Then, we leverage the body motion embedding as a \emph{query} to retrieve the most consistent single hand motion embedding $F_{\mathcal{SHM}}\in \mathcal{R}^{T\times 1}$ stored in a temporal memory bank. The initialization and reading processes of the temporal memory bank are similar to SRM. 
We utilize the body motion embedding to \emph{update} the retrieved slot of the temporal memory bank in the same manner as SRM.

Afterward, we utilize the retrieved hand motion embedding to boost the body-hand temporal consistency. Specifically, we leverage the retrieved temporal embedding to represent the temporal correlation of hand gesture changes in a sequence.
Then, the temporal-consistent enhanced sequential hand gestures are attained by:
$F_{\mathcal{SH}} \!=\!  F_{\mathcal{SH}}  + F_{\mathcal{SH}} \odot softmax\left ( F_{\mathcal{SHM}} \right )$, where $\odot$ means dot product.

\vspace{0.5em}
\noindent{\textbf{Transformer-based Backbone:}}
As illustrated in Fig. \ref{fig:figure2} (stage one), to further enhance the body-hand interaction, we construct the bilateral hand disentanglement branches that interact with a body-specific branch to predict 3D hands. In particular, we introduce a body-specific transformer to generate body features as query $Q$. Then, through 3 times Multi-Head Attention (MHA) \cite{vaswani2017attention}, we use $Q$ to match the key features $K$ and value features $V$ in each hand transformer branch, respectively.

Furthermore, we feed the concatenation of bilateral hand features into an MLP, thus obtaining the merged two-hand features. Similarly, we conduct 3 times MHA in the transformer-based decoder. This encourages the merged two-hand features to be more temporal consistent with body dynamics, holistically.
At the end of the decoder, we adopt two fully-connected layers to obtain the 3D two hands sequence. Additionally, similar to \cite{Ng_2021_CVPR}, we employ a motion discriminator to ensure the natural and continuous initial prediction at stage one.

\subsection{Stage Two: Diverse Sampling}\label{Diverse_Sampling}

We leverage a Prototypical-Memory Sampling Strategy (PSS) to achieve hands diversification based on our initial predictions from stage one. Specifically, we first construct a prototypical memory to store the realistic two hands prototypes. Then, we obtain the retrieved realistic prototype representation based on the initial prediction at stage one. Meanwhile, as shown in Fig. \ref{fig:figure2} (stage two), we propose a sampling header to obtain the hand-informative perturbation via gradient-based Markov Chain Monte Carlo (MCMC) sampling. Finally, we concatenate the retrieved realistic hand prototype representation and perturbation to generate the diverse 3D two hands. 

\vspace{0.5em}
\noindent{\textbf{Hands Prototypical Memory Construction:}}
Inspired by \cite{zhang2021learning}, we reformulate the two hands diversification as a generation task. Instead of directly sampling the prior of the perturbation from the isotropic Gaussian distribution \cite{kingma2013auto,goodfellow2020generative, han2017alternating, sohn2015learning}, we assume the sampling space follows a learnable realistic prior distribution \cite{pang2020learning, pang2021latent, zhang2021learning}. Since the existing 3D hand prediction dataset is noisy due to the automated annotation process, we propose a prototypical memory bank to store the realistic hand prototype representations encoded from real 3D hands. These 3D hands are captured from a studio-based mocap-captured dataset, named BEAT \cite{liu2022beat}. The reading and updating strategies of hands prototypical memory are the same as TMM. 

\vspace{0.5em}
\noindent{\textbf{Gradient-based MCMC Sampling:}}
We exploit the retrieved hand prototype representation concatenated with sampled perturbation vector $w$ to predict the diverse hand gestures. Given the observed 3D two hands $h\in \mathbb{R} ^{90}$, the $\tilde{h}$ is expressed as:
\begin{equation}
\tilde{h} = R_{\theta } (h,  w) + \epsilon, w\sim p_{\alpha}(w), \epsilon\sim \mathcal{N}(0,\sigma_{\epsilon }^{2}I), 
\label{eq3}
\end{equation}
where $ p_{\alpha}(w)$ is the prior distribution of perturbation with parameters $\alpha$, and $\epsilon\sim \mathcal{N}(0,\sigma_{\epsilon }^{2}I)$ is the observation noise of hand gestures with $\sigma_{\epsilon }$ being given, $R_{\theta }$ is the generation model with parameters $\theta$. 

Inspired by \cite{han2017alternating}, we leverage an MLP-based sampling header $S_{\alpha }(w)$ to model the diversification sampling process, as illustrated in Fig. \ref{fig:figure2} (stage two). The prior distribution of perturbation is initialized from an isotropic Gaussian reference distribution, expressed as:
\begin{equation}
p_{\alpha }(w) \propto exp\left [ -S_{\alpha }(w)-\frac{1}{2\sigma _{w}^{2} }\left \|  w\right \| ^{2}    \right ], 
\label{eq4}
\end{equation}
where the $M_{\alpha }(w) = S_{\alpha }(w)+\frac{1}{2\sigma _{w}^{2} }\left \| w\right \|^{2}$ is defined as the whole sampling function, and $\alpha$ is the learnable parameters of sampling header. The hyperparameter $\sigma _{w}$ denotes the standard deviation.
For notation simplicity, let $\beta = \left \{ \theta , \alpha \right \} $. For the $i$ th sample in a training mini-batch with size $n$, the log-likelihood function of $\beta$ is defined as:
\begin{align}
L(\beta ) = \sum_{i=1}^{n} log\left [ \int p_{\alpha }(w_{i})p_{\theta}(\tilde{h}_{i}  | h_{i}, w_{i} )dw_{i}     \right ].
\label{eq5}
\end{align}
Thus the gradient of $L(\beta )$ is computed as:
\begin{align}
\bigtriangledown L(\beta ) = E_{p_{\beta (w|\tilde{h} ,h) } }\left [ \bigtriangledown _{\alpha  } log p_{\alpha } (w) + \bigtriangledown _{\theta }log p_{\theta  } (\tilde{h}|h, w)\right ] .
\label{eq6}
\end{align}


The intractable expectation term $E_{p}(\cdot )$ in Eq. \eqref{eq6} is approximately solved by a gradient-based MCMC (Langevin  dynamics) \cite{neal2011mcmc, dubey2016variance}. 
In particular, the initial state of perturbation $p_{0}(w) $ is sampled from the Gaussian distribution. 
Through MCMC, the perturbation from the prior distribution and posterior distribution are represented as $\left \{ w_{i}^{-}  \right \}$ and $\left \{ w_{i}^{+} \right \}$, respectively. The detailed calculation process is in \cite{neal2011mcmc}.
Therefore, the gradients of the generation model and sampling header are updated by :
\begin{align}
\bigtriangledown_{\theta } R_{\theta }\!&=\! \frac{1}{n} \sum_{i  = 1}^{n} \left [ \frac{1}{\sigma _{\epsilon }^{2} } (\tilde{h}_{i}  -R_{\theta } (h_{i} , w_{i}^{+} ))\bigtriangledown _{\theta } R_{\theta }(h_{i} ,w_{i}^{+} ) \right ],\\
\bigtriangledown_{\alpha } S_{\alpha } \!&=\! \frac{1}{n} \sum_{i = 1}^{n} \left [ \bigtriangledown _{\alpha }S_{\alpha } (w_{i}^{-} ) \right ]\!-\!\frac{1}{n} \sum_{i = 1}^{n} \left [ \bigtriangledown _{\alpha }S_{\alpha } (w_{i}^{+} ) \right ].
\label{eq7}
\end{align}

\begin{table*}[t]
\centering
\caption{Comparison with the state-of-the-art methods on \textbf{B2H dataset} and \textbf{our proposed TED Hands dataset} of the first stage initial prediction. $\downarrow$ indicates the lower the better.}
\vspace{-0.5em}
\label{tab:table1}
\setlength{\tabcolsep}{5 mm}
\footnotesize
\begin{tabular}{lcccccc}
\toprule
\multicolumn{1}{c}{\multirow{2}{*}{Methods}} & \multicolumn{3}{c}{B2H Dataset \cite{Ng_2021_CVPR}}                      & \multicolumn{3}{c}{TED Hands Dataset}               \\ \cline{2-7} 
\multicolumn{1}{c}{}                         & L2$\downarrow$              & FHD$\downarrow$             & MPJRE$\downarrow$            & L2$\downarrow$              & FHD$\downarrow$             & MPJRE$\downarrow$           \\ \midrule
Body2Hands \cite{Ng_2021_CVPR}\textcolor[HTML]{C0C0C0}{$_{CVPR'21}$}                                   & 3.760          & 1.103          & 13.148          & 2.551          & 1.174          & 11.371         \\
MRT \cite{wang2021multi}\textcolor[HTML]{C0C0C0}{$_{NeurIPS'21}$}  & 3.634          & 0.973          & 12.576          & 2.325          & 0.877          & 10.314         \\
BTM \cite{guo_2023_wacv}\textcolor[HTML]{C0C0C0}{$_{WACV'23}$} & 3.657          & 1.263          & 12.696          & 2.350          & 1.111          & 10.440         \\
LTD \cite{mao2019learning}\textcolor[HTML]{C0C0C0}{$_{ICCV'19}$}& 3.755          & 1.073          & 13.315          & 2.482          & 1.367          & 11.078         \\
MotionMixer  \cite{ijcai2022p111}\textcolor[HTML]{C0C0C0}{$_{IJCAI'22}$} & 3.616          & 0.986          & 12.702          & 2.324          & 0.910          & 10.427         \\
SPGSN  \cite{li2022skeleton}\textcolor[HTML]{C0C0C0}{$_{ECCV'22}$}  & 3.656          & 0.962          & 12.757          & 2.435          & 0.990          & 10.887         \\ \midrule
\rowcolor[HTML]{EFEFEF} 
\textbf{Ours (stage one)}                      & \textbf{3.420} & \textbf{0.471} & \textbf{11.406} & \textbf{2.037} & \textbf{0.258} & \textbf{8.888} \\ \bottomrule
\end{tabular}
\vspace{-1em}
\end{table*}

\subsection{Objective Functions}\label{Training_Objectives}
In stage one, we apply the following loss functions to constrain the initial prediction process of hand gestures.

\vspace{0.5em}
\noindent{\textbf{Reconstruction Loss:}}
We use the ground truth to supervise the predicted two hands as $\mathcal{L}_{rec} \!=\! \left \| \mathcal{H}\!-\! \tilde{\mathcal{H}}\right \| _{1}$, 
where $\tilde{\mathcal{H}}$ denotes generated hands.

\vspace{0.5em}
\noindent{\textbf{Perceptual Loss:}}
Inspired by \cite{johnson2016perceptual}, we leverage the studio-based 3D hands dataset BEAT to pre-train an MLP-based autoencoder as a feature extractor $\phi\left ( \cdot  \right ) $. We utilize the $L1$ loss to constrain the difference between the extracted features of $\mathcal{H}$ and $\tilde{\mathcal{H}}$, as formulated: $\mathcal{L}_{perc} = \left \| \phi (\mathcal{H}) - \phi (\tilde{\mathcal{H}})\right \| _{1}$.

\vspace{0.5em}
\noindent{\textbf{Adversarial Learning:}}
Following the configuration of \cite{Ng_2021_CVPR}, we employ the adversarial training loss as:
\begin{align}
\mathcal{L}_{adv} = \mathbb{E} _{\mathcal{H}} \left [ log D(\mathcal{H}) \right ] + \mathbb{E} _{\mathcal{B}} \left [ log(1- (G(\mathcal{B})) \right ],
\end{align}
where $D$ denotes the discriminator and $G$ means hands generator. Finally, the overall objective in stage one is:
\begin{align}
\!\!\min_{G}\max_{D}\mathcal{L}_{total}^{stage1} \!=\!  \mathcal{L} _{rec} +  \mathcal{L} _{adv} +\mathcal{L} _{perc} + 0.5 \mathcal{L} _{dis},
\label{eq11}
\end{align}
where $\mathcal{L} _{dis}$ denotes disentanglement constraint in Eq. (\eqref{Eq1}). \\
\indent In the second stage, the overall objective to train our model is defined as follows: 
\begin{align}
\mathcal{L}_{total}^{stage2} = \left \| h - R_{\theta } (h,  w^{+} ) \right \| _{1} ,
\label{eq12}
\end{align}
where $w^{+}$ is the posterior distribution of the perturbation. We conduct the frame-level diversification based on the initial prediction at stage one. To ensure temporal smoothness, we further apply temporal smoothness \cite{qvi_nips19} to the diversified 3D hand gestures as a post processing.


\begin{table}[t]
\centering
\caption{Sampling experiment results on \textbf{B2H dataset}. $\downarrow$ indicates the lower the better, and $\uparrow$ indicates the higher the better. $\pm$ means 95\% confidence interval.} 
\vspace{-0.5em}
\label{tab:table2}
\footnotesize
\setlength{\tabcolsep}{5mm}{%
\renewcommand{\arraystretch}{0.98} 
\begin{tabular}{lcc}
\toprule
\multicolumn{1}{c}{Methods} & FHD$\downarrow$    & Diversity$\uparrow$ \\ \midrule
Body2hands \cite{Ng_2021_CVPR} w/ PSS & 2.504 & $0.497^{\pm 0.028}$ \\
MRT \cite{wang2021multi} w/ PSS & 1.982 & $0.698^{\pm 0.031}$ \\
BTM \cite{guo_2023_wacv} w/ PSS & 2.379 & $0.703^{\pm 0.038}$ \\
LTD \cite{mao2019learning} w/ PSS & 1.893 & $0.577^{\pm 0.028}$ \\
MotionMixer \cite{ijcai2022p111} w/ PSS & 2.169 & $0.869^{\pm 0.055}$ \\
SPGSN \cite{li2022skeleton} w/ PSS & 2.004 & $0.874^{\pm 0.046}$ \\ 
\rowcolor[HTML]{f7f7f7} 
Ours w/ VAE \cite{sohn2015learning} & 1.617 & 1.001$^{\pm 0.061}$ \\
\midrule
\rowcolor[HTML]{EFEFEF} 
\textbf{Ours (stage two)} & \textbf{1.239} & \textbf{1.324$^{\pm 0.066}$} \\\bottomrule
\end{tabular}%
}
\end{table}


\section{Experiments}

\subsection{Datasets and Experimental Setting}
\noindent{\textbf{B2H Dataset:}}
The B2H dataset \cite{Ng_2021_CVPR} is automatically annotated by a monocular-based 3D pose estimation model \cite{xiang2019monocular}, including the gestures of 8 speakers from in-the-wild talk show videos. In the original B2H dataset, there are more than 139K sequences. Each sequence includes 64 frames, with 32 frames overlapping between every two sequences.
However, due to the noisiness and unavailable videos, we obtain 120,188 sequences finally. In our work, we adopt the same division criteria as \cite{Ng_2021_CVPR} to split the training and testing dataset.

\vspace{0.5em}
\noindent{\textbf{TED Hands Dataset:} }
Due to the limited avatar identities in the existing 3D hand prediction dataset, we newly collect a large-scale 3D hands dataset (dubbed TED Hands) based on the TED Gesture \cite{yoon2020speech, yoon2019robots}. The original TED Gesture dataset only contains 10 upper body joints without elaborate fingers of two hands. Hence, we leverage a state-of-the-art 3D pose estimator FrankMocap \cite{Rong_2021_ICCV} to extract the whole body-hand joints as pseudo ground truth from the video links provided by TED Gesture.  Concretely, each joint of our dataset is normalized as the 3D axis-angle representation in a fixed kinematic structure \cite{romero2017embodied}. In this fashion, our dataset is invariant to root joint motion and body shape. Finally, we acquired the 134,456 sequences from 1,755 different speakers, with 99.6 hours of TED talking speeches. Each sequence includes 64 frames without overlap. We randomly spit the dataset into the 70\% training set, 10\% validation set, and 20\% testing set. 
Please refer to the supplementary material for more details.

\vspace{0.5em}
\noindent{\textbf{Implementation Details:}}
Our model is implemented on the PyTorch platform with 2 NVIDIA RTX 2080Ti GPUs. We train the model utilizing Adam optimizer with an initial learning rate of 0.003. The whole training takes 30 epochs with a batch size of 64. We set $\sigma_{\epsilon } = 1$ in Eq. (\ref{eq3}), and $\sigma_{w } = 1$ in Eq. (\ref{eq4}).

\begin{table}[t]
\centering
\caption{Sampling experiment results on \textbf{our proposed TED Hands dataset}. $\downarrow$ indicates the lower the better, and $\uparrow$ indicates the higher the better. $\pm$ means 95\% confidence interval.}
\vspace{-0.5em}
\label{tab:table3}
\footnotesize
\setlength{\tabcolsep}{5mm}{%
\renewcommand{\arraystretch}{0.98} 
\begin{tabular}{lcc}
\toprule
\multicolumn{1}{c}{Methods} & FHD$\downarrow$    & Diversity$\uparrow$ \\ \midrule
Body2hands \cite{Ng_2021_CVPR} w/ PSS & 2.005 & $2.705^{\pm 0.153}$ \\
MRT \cite{wang2021multi} w/ PSS & 1.188 & $4.954^{\pm 0.190}$ \\
BTM \cite{guo_2023_wacv} w/ PSS & 1.224 & $3.632^{\pm 0.150}$ \\
LTD \cite{mao2019learning} w/ PSS & 1.385 & $2.851^{\pm 0.101}$ \\
MotionMixer \cite{ijcai2022p111} w/ PSS & 1.613 & $3.487^{\pm 0.208}$ \\
SPGSN \cite{li2022skeleton} w/ PSS & 1.565 & $3.812^{\pm 0.219}$ \\ 
\rowcolor[HTML]{f7f7f7} 
Ours w/ VAE \cite{sohn2015learning} & 1.033 & 4.018$^{\pm 0.190}$ \\
\midrule
\rowcolor[HTML]{EFEFEF} 
\textbf{Ours (stage two)} & \textbf{0.938} & \textbf{6.426$^{\pm 0.300}$} \\ \bottomrule
\end{tabular}%
}
\end{table}

\vspace{0.5em}
\noindent{\textbf{Evaluation Metrics:}}
To fully evaluate the naturalness of initial predictions at stage one and the diversity of sampled hand gestures at stage two, we employ various metrics:
\begin{itemize}[leftmargin=*]
    \vspace{-0.6em}
    \item  \textbf{$L2$ Distance}: Distance between the generated whole two hand poses and pseudo ground truth. 
    \vspace{-0.6em}
    \item \textbf{FHD}: Inspired by FGD \cite{yoon2020speech, yoon2019robots}, we propose a hand-specific metric (named FHD) to measure the Fréchet distance between the features of generated hands and pseudo ground truth. The feature extractors of FHD are pre-trained by B2H and TED Hands datasets, respectively.
    \vspace{-0.6em}
    \item \textbf{MPJRE}: We leverage the Mean Per Joint Rotation Error $\left [ \circ  \right ] $ (MPJRE) \cite{jiang2022avatarposer} to measure the absolute distance between predicted 3D representation joints and pseudo ground truth.  
    \vspace{-0.6em}
    \item \textbf{Diversity}: To verify the diversity of sampled hand gestures, we calculate the average feature distance between 500 random combined sequential pairs \cite{liu2022learning, liang2022seeg}. The feature extractor is the same as FHD.
\end{itemize}

\subsection{Quantitative Results}

\begin{table*}[ht]
\centering
\caption{Ablation study on different loss functions and different components of the proposed method. $\downarrow$ indicates the lower the better. $\checkmark$ indicates the employment of a certain module or loss function.  \# means the temporal-motion memory module is adopted before the spatial-residual memory module. $\dagger$ denotes the weights of the two hands transformers are shared. }
\vspace{-1em}
\label{tab:table4}
\footnotesize
\setlength{\tabcolsep}{2.5mm}{%
\renewcommand{\arraystretch}{0.95} 
\begin{tabular}{cccccc|cccccc}
\toprule
\multicolumn{6}{c|}{Model Variations} & \multicolumn{3}{c}{B2H Dataset \cite{Ng_2021_CVPR}} & \multicolumn{3}{c}{TED Hands Dataset} \\ \midrule
Baseline & Full Model & $\mathcal{L} _{dis}$ & $\mathcal{L}_{perc}$  & SRM & TMM & L2 $\downarrow$ & FHD $\downarrow$ & MPJRE $\downarrow$ & L2 $\downarrow$ & FHD $\downarrow$ & MPJRE $\downarrow$ \\ \midrule
\checkmark &  &  &  &  &  & 3.650 & 1.091 & 12.690 & 2.366 & 0.977 & 10.531 \\
 & \checkmark &  &  &  &  & 3.604 & 0.773 & 12.453 & 2.303 & 0.731 & 10.250 \\
 & \checkmark &\checkmark  & &  &  & 3.552 & 0.682 & 12.037 & 2.216 & 0.595 & 9.786 \\
  & \checkmark &  & \checkmark &  &  & 3.563 & 0.657 & 12.085 & 2.261 & 0.699 & 10.026 \\
 & \checkmark & \checkmark & \checkmark &  &  & 3.504 & 0.640 & 11.828 & 2.161 & 0.495 & 9.508 \\
 & \checkmark & \checkmark & \checkmark & \checkmark &  & 3.457 & 0.580 & 11.687 & 2.100 & 0.393 & 9.209 \\
 \rowcolor[HTML]{f7f7f7} 
 \# & \checkmark & \checkmark & \checkmark & \checkmark & \checkmark & 3.484 & 0.614 & 11.758 & 2.124 & 0.463 & 9.311 \\
 \rowcolor[HTML]{f7f7f7} 
 $\dagger$ & \checkmark & \checkmark & \checkmark & \checkmark & \checkmark & 3.476 & 0.598 & 11.748 & 2.156 & 0.380 & 9.467 \\  \midrule
\rowcolor[HTML]{ededed} 
  & \checkmark & \checkmark & \checkmark & \checkmark & \checkmark & \textbf{3.420} & \textbf{0.471} & \textbf{11.406} & \textbf{2.037} & \textbf{0.258} & \textbf{8.888} \\ \bottomrule
\end{tabular}%
}
\end{table*}

\noindent{\textbf{Comparisons with the state-of-the-art:}}
We compare our method with the pioneering 3D hands prediction work, named Body2hands \cite{Ng_2021_CVPR}. Moreover, to fully verify the superiority of our method, we implement various state-of-the-art (SOTA) human motion prediction models: MTR\cite{wang2021multi}, BTM\cite{guo_2023_wacv}, LTD\cite{mao2019learning}, MotionMixer\cite{ijcai2022p111}, and SPGSN \cite{li2022skeleton}.
For fair comparisons, all methods are implemented by source codes or pre-trained models released by authors. Since the above counterparts are designed without the diversification setting, the comparison is divided into two parts. 

In the first part, we leverage our stage one initial prediction to compare with other competitors. We adopt the $L2$ distance, FHD, and MPJRE to provide a well-rounded view of comparisons. As reported in Tab. \ref{tab:table1}, our method outperforms all the counterparts by a large margin. Remarkably, our method achieves the lowest FHD and MPJRE sores on both datasets. Concretely, on FHD scores of the TED Hands dataset, our method performs even 70.6\% lower than the optimal results of other competitors (\emph{e.g.,} $(0.877-0.258) / 0.877 \approx 70.6\%$). 

To demonstrate the effectiveness of the diversification process in our method, we compare the diversified results based on the various predictions from above mentioned competitive methods. 
Meanwhile, we replace our MCMC-based sampling strategy with CVAE \cite{sohn2015learning} to diversify our initial prediction at stage one. 
We exploit the FHD and Diversity metrics in the second part. As shown in Tab. \ref{tab:table2} and Tab. \ref{tab:table3}, our method achieves superior performance from the perspective of diversity. Although the FHD is slightly worse than stage one due to the sampled perturbation, our method still achieves the optimal result. Compared with CVAE, our method effectively avoids the potential posterior collapse problem \cite{he2018lagging}. Moreover, we observe that the diversity scores on our TED Hands dataset are much higher than the B2H dataset with all the counterparts. It applies that our newly collected TED Hands dataset includes more diverse hand gestures. 

\vspace{0.5em}
\noindent{\textbf{Ablation Study:}}
We further conduct the ablation study based on stage one to verify the effectiveness of each module and different loss functions, as indicated in Tab. \ref{tab:table4}.  
The \textbf{\emph{baseline model}} is implemented by a simple transformer-based pipeline without bilateral hand branches. Our \textbf{\emph{full model}} adopts the bilateral hand branches interacting with a body-specific transformer to produce the 3D hand gestures. 

As shown in Tab. \ref{tab:table4}, all combinations of the different loss functions and components have positive impacts on 3D hand predictions. The most complete combination attains the best results.
Even though we directly construct the bilateral branches without disentanglement constraints (\emph{i.e.,} $\mathcal{L}_{dis}$) as \textbf{\emph{full model}}, our method achieves comparable performance against the simple \textbf{\emph{baseline}}. This indicates the global body-hand interaction are effectively modeled by bilateral hand transformers and a body-specific transformer. Then, we observe that after adding the \textbf{\emph{loss functions}} $\mathcal{L}_{dis}$ and $\mathcal{L}_{perc}$, the performance of our method is significantly improved, especially in MPJRE scores on both datasets (\emph{e.g.,} $10.250\to9.508$ on TED Hands). 

To verify the effectiveness of \textbf{\emph{SRM}} and \textbf{\emph{TMM}}, we conduct plenty of experiments on both datasets. As reported in Tab. \ref{fig:figure4}, when we first employ the SRM, our method achieves the lower $L2$ distance and FHD scores on both datasets. Then, after adding TMM, our method achieves the best performance. This result proves that TMM effectively models the body-hand temporal association. Besides, we find that when TMM is adopted before SRM, the model performance decreases slightly (\textbf{\emph{the third row from bottom to top}} in Tab. \ref{tab:table4}). This corresponds to both TMM and transformer-based backbone (\emph{i.e.} body-transformer, bilateral hand branches) focus on modeling the body-hand temporal correlation. Both of them encourage the two hands to be more temporal consistent with body dynamics. 
Employing the SRM between them leads to worse spatial-temporal modeling for body-hand interaction.  

Moreover, we conduct experiments that the parameter weights of bilateral hand transformers are shared. As shown in Tab. \ref{tab:table4} (\textbf{\emph{the second row from bottom to top}}), the performance is decreased due to the spatial-temporal interaction between the body and single-hand-specific branch being merged here. This supports our key technical insight on bilateral hand disentanglement.


\subsection{Qualitative Evaluation}

\begin{figure*}[t]
\begin{center}
\includegraphics[width=0.95\linewidth]{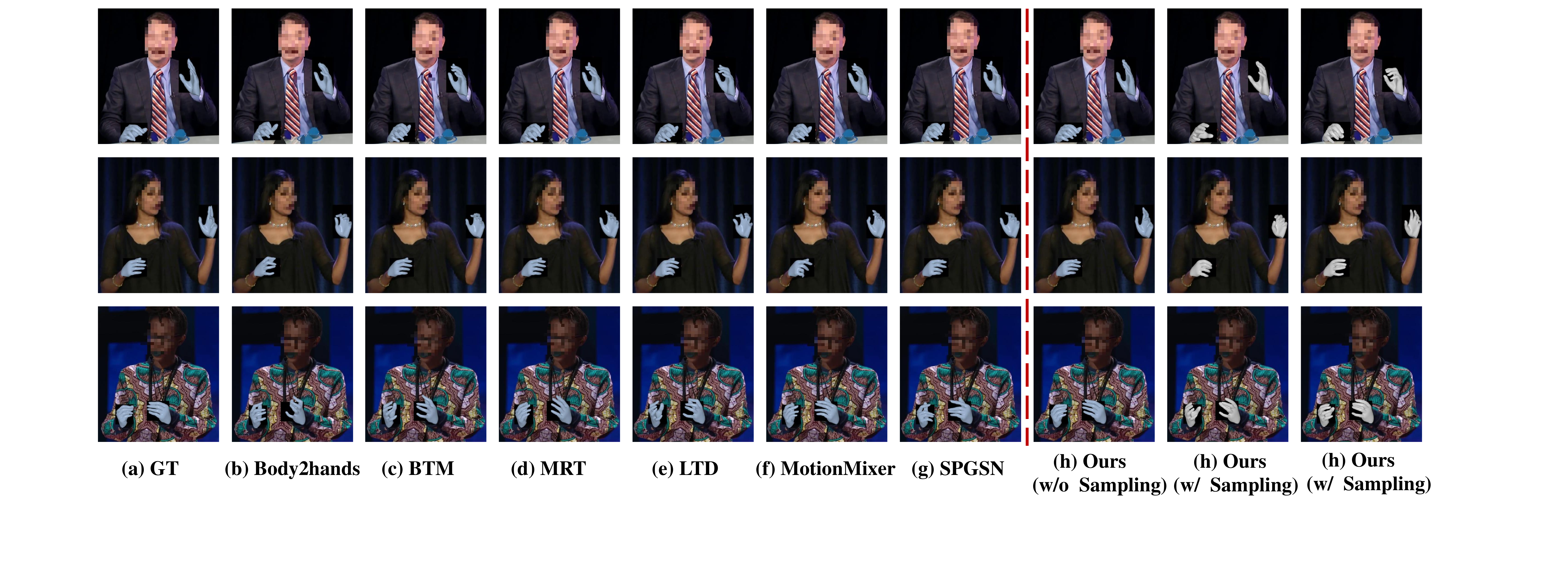}
\end{center}
\vspace{-6.0 mm}
\caption{ Visualization of our predicted 3D hand gestures against various state-of-the-art methods \cite{Ng_2021_CVPR,guo_2023_wacv,wang2021multi,mao2019learning,ijcai2022p111, li2022skeleton}. From top to bottom, samples of the first row are from the B2H dataset, and the others are from our TED Hands dataset. Best view on screen.
}
\label{fig:figure4}

\end{figure*}
\begin{figure}[t]
\begin{center}
\includegraphics[width=0.95 \linewidth]{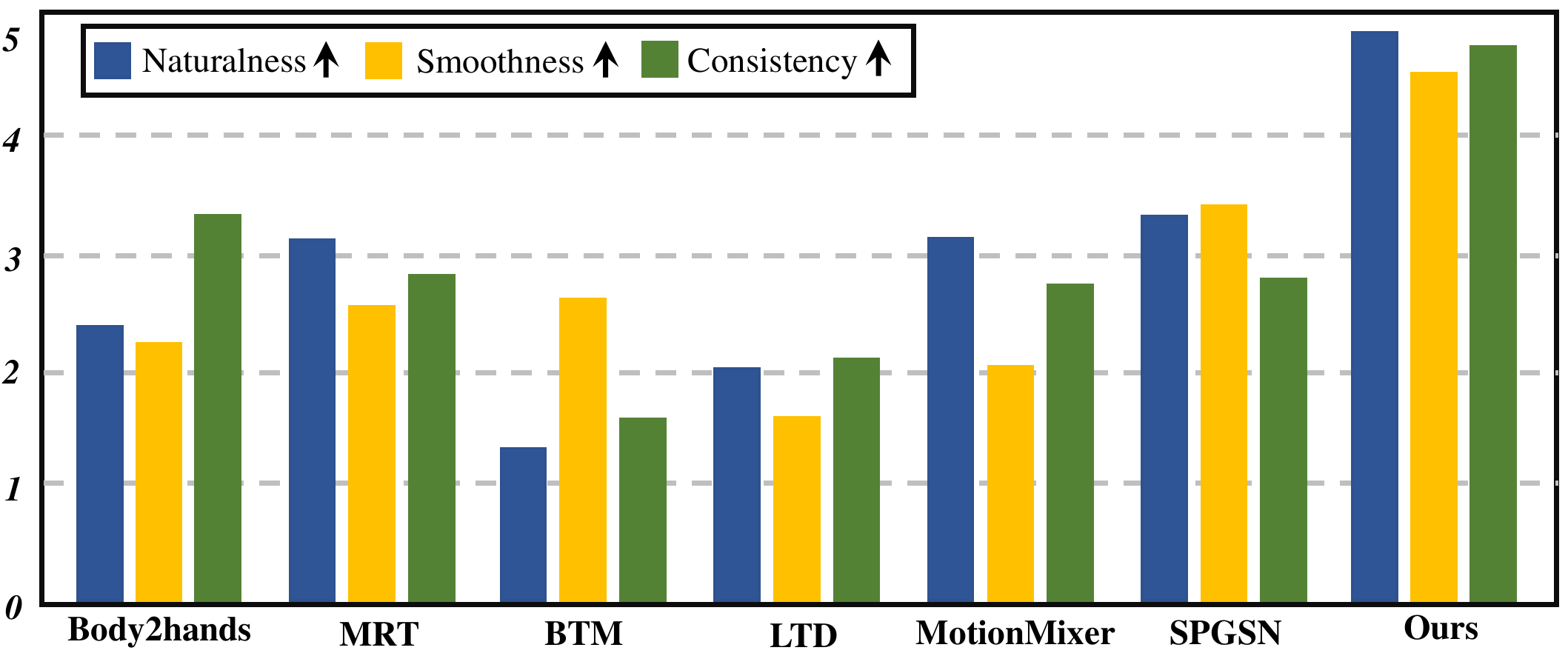}
\end{center}
\vspace{-1.5em}
\caption{ User study on the motion naturalness, smoothness, and body-hand temporal consistency. $\uparrow$ indicates the higher the better. }
\vspace{-0.5em}
\label{fig:user_study}
\end{figure}

\noindent{\textbf{User Study:}}
To further analyze the visual quality of the predicted 3D hand gestures by different methods, we recruit 15 volunteers to participate in the user study. We take the naturalness, smoothness, and body-hand temporal consistency as three evaluation perceptions and the range of grades is from 0 to 5 (the higher, the better). The statistical results are reported in Fig. \ref{fig:user_study}. Our method shows the best performance in all three metrics. Especially in the naturalness and body-hand temporal consistency, our method achieves remarkable advantages among all the methods due to the body-hand interaction modeling by SRM and TMM.

\vspace{0.5em}
\noindent{\textbf{Visualization:}}
To fully verify the performance of our method, we show the visualization of generated 3D hands from the first and second stages, respectively. As depicted in Fig. \ref{fig:figure4}, we compare our initial predictions (from stage one) with various SOTA approaches. The avatars are rendered from ground truth speakers in B2H and TED Hands datasets.
We observe that Body2hands, BTM, and LTD generate unreasonable hand gestures (\emph{e.g.,} in the third row of Body2hands, left-hand gestures of avatar).  MRT, MotionMixer, and SPGSN predict the natural results, but the finger gestures are less consistent with the upper body dynamics. In contrast, our method successfully predicts the perceptually convincing 3D hand gestures from the body dynamics. This supports our key insight to build the bilateral hand disentanglement branches since the asymmetric motions of two hands. We further visualize the diversification results generated by our method at stage two. As shown in Fig. \ref{fig:figure1} and Fig. \ref{fig:figure4}, with the same upper body dynamics as input, our method obtains the natural and diverse 3D hand gestures on B2H and TED Hands datasets, respectively. Please refer to the supplementary material for more visualization results.

\section{Conclusion}

In this paper, we present a novel bilateral hand disentanglement based two-stage method towards natural and diverse 3D hand prediction from body dynamics. In the first stage, upon the bilateral hand branches, we fully take advantage of the spatial-temporal body-hand interaction for generating natural initial 3D hand gestures. Then, in the second stage, we further diversify our natural initial 3D hand gestures by gradient-based markov chain monte carlo sampling. 
Extensive experiments conducted on B2H and our newly collected TED Hands datasets demonstrate the superiority of our method. 
As our method aims at generating diverse 3D avatar hand postures, in future work, we will investigate sampling diverse 3D hand gestures with temporal smoothness, instead of frame-wise sampling.

\vspace{0.5em}
\noindent{\textbf{Acknowledgements.}} This research is funded in part by ARC-Discovery grant (DP220100800 to XY) and ARC-DECRA grant (DE230100477 to XY). 

\newpage

{\small
\bibliographystyle{ieee_fullname}
\bibliography{PaperForReview}
}

\clearpage



\section*{Appendix}

\section*{A. Architecture Details}
\paragraph{Body Encoder.}
The body encoder aims to encode the input upper body skeletons into the body features via an MLP-based architecture. The channel dimension of body features is $C=128$ in practice. 

\paragraph{Bilateral Hand Disentanglement Transformers.}
We design bilateral hand transformers that interacted with a body-specific transformer. Concretely, we leverage the body features $Q$ to match the key features $K$ and value features $V$ in a single-hand-specific transformer via 3 times Multi-Head Attention (MHA) \cite{vaswani2017attention}, expressed as:
\begin{align}
MultiHead_{F_{\mathcal{B}\to \mathcal{SH}}} (Q, K, V) = softmax(\frac{QK }{\sqrt{d} } )V,
\end{align}
where $d$ is a normalization constant. 

\begin{table*}[t]
\centering
\caption{Statistics of the B2H, TED Gestures, and our newly collected TED Hands datasets.}
\setlength{\tabcolsep}{2 mm}
\footnotesize
\begin{tabular}{cccccc}
\hline
Dataset & Speaker Identities & Interest Shots Length & Sequence Numbers & \begin{tabular}[c]{@{}c@{}}Frame Numbers\\ in a Sequence\end{tabular}  & \begin{tabular}[c]{@{}c@{}}Speaker Identities\\ in a Sequence\end{tabular} \\ \hline
B2H \cite{Ng_2021_CVPR} & 8 & 71.2h & 120,188 & 64 & Only one \\
TED Gestures \cite{yoon2020speech, yoon2019robots} & 1,766 & 106.1h & 252,109 & 34 & Multi-identities \\
TED Hands & 1,755 & 99.6h & 134,456 & 64 & Only one \\ \hline
\end{tabular}%
\label{tab:statistics}
\end{table*}

\section*{B. More Details about MCMC Sampling}
Inspired by \cite{han2017alternating}, we leverage an MLP-based sampling header $S_{\alpha }(w)$ to model the diversification sampling process, where $w$ indicates the perturbation vector and $\alpha$ is the parameter of sampling header. The prior distribution of perturbation is initialized from an isotropic Gaussian reference distribution, expressed as:
\begin{align}
p_{0}(w)=\mathcal{N}(0,\sigma_{w }^{2}I),
\end{align}
where the hyperparameter $\sigma _{w}$ denotes the standard deviation. The whole sampling process is formulated as:
\begin{equation}
p_{\alpha }(w) \propto exp\left [ -S_{\alpha }(w)-\frac{1}{2\sigma _{w}^{2} }\left \|  w\right \| ^{2}    \right ], 
\label{eq13}
\end{equation}
where the $M_{\alpha }(w) = S_{\alpha }(w)+\frac{1}{2\sigma _{w}^{2} }\left \| w\right \|^{2}$ is defined as the whole sampling function, and $\alpha$ is the learnable parameters of sampling header. For notation simplicity, let $\beta = \left \{ \theta , \alpha \right \} $. For the $i$ th sample in a training mini-batch with size $n$, the log-likelihood function of $\beta$ is defined as:
\begin{align}
L(\beta ) = \sum_{i=1}^{n} log\left [ \int p_{\alpha }(w_{i})p_{\theta}(\tilde{h}_{i}  | h_{i}, w_{i} )dw_{i}     \right ].
\label{eq14}
\end{align}
Thus the gradient of $L(\beta )$ is computed as:
\begin{align}
\bigtriangledown L(\beta ) = E_{p_{\beta (w|\tilde{h} ,h) } }\left [ \bigtriangledown _{\alpha  } log p_{\alpha } (w) + \bigtriangledown _{\theta }log p_{\theta  } (\tilde{h}|h, w)\right ] .
\label{eq15}
\end{align}
We decompose the $\bigtriangledown L(\beta )$ into two parts. 
The first part is the gradient for the sampling header with parameter $\alpha$:
\begin{align}
E_{p_{\beta (w|\tilde{h} ,h) } }\left [ \bigtriangledown _{\alpha } log p_{\alpha } (w) \right ] & = E_{p_{\alpha } (w)}\left [ \bigtriangledown _{\alpha }S_{\alpha } (w)  \right] & 
\notag
\\&-E_{p_{\beta (w|\tilde{h} ,h) } }\left [ \bigtriangledown _{\alpha }S_{\alpha } (w)  \right ].  
\label{eq16}
\end{align}
The second part is the gradient for the hand generation model with parameter $\theta$:
\begin{align}
&E_{p_{\beta (w|\tilde{h} ,h) } }\left [ \bigtriangledown _{\theta  } log p_{\theta  } (\tilde{h}|h, w) \right ] 
\notag
\\& = E_{p_{\beta} (\tilde{h}|h, w)}\left [ \frac{1}{\sigma _{\epsilon }^{2} } (\tilde{h} -R_{\theta } (h, w))\bigtriangledown _{\theta } R_{\theta }(h,w)  \right ].
\label{eq17}
\end{align}
In practice, the terms $\bigtriangledown _{\alpha }S_{\alpha } (w)$ in Eq. \eqref{eq16} and $\bigtriangledown _{\theta } R_{\theta }(h,w)$ in Eq. \eqref{eq17} are directly computed by back-propagation.
The intractable expectation terms $E_{p}(\cdot )$ in Eq. \eqref{eq16} and Eq. \eqref{eq17} are approximately solved by a gradient-based MCMC (Langevin  dynamics) \cite{neal2011mcmc}. Specifically, the perturbation is obtained from the MLP-based prior sampling process $M_{\alpha }(w)$, by iterating:
\begin{align}
w^{l+1}  = &w^{l} -\delta \bigtriangledown _{w}M_{\alpha} (w^{l} )+\sqrt{2\delta}e^{l}, 
\notag
\\& w_{0} \sim p_{0}(w), e^{l}\sim \mathcal{N} (0,I),
\label{er18}
\end{align}
where $l$ denotes the $l$ th iteration state, and $\delta $ is the step size of Langevin sampling. Meanwhile, the posterior distribution $p_{\beta  } (w|\tilde{h}, h)$ of the perturbation is computed by iterating:\vspace{-2mm}
\begin{align}
w^{l+1} &= w^{l} -
\notag
\\&\delta \left [  \bigtriangledown _{w}M_{\alpha} (w^{l} ) - \frac{1}{\sigma _{\epsilon }^{2} } (\tilde{h} -R_{\theta } (h, w^{l} ))\bigtriangledown _{w} R_{\theta }(h,w^{l} )\right ]
\notag
\\&+\sqrt{2\delta}e^{l}, w_{0} \sim p_{0}(w), e^{l}\sim \mathcal{N} (0,I).
\label{er19}
\end{align}
In the experiments, we set the total iteration state as 6, and the Langevin step sizes of the prior and posterior are 0.4 and 0.1, respectively.

\section*{C. More Details about Datasets}
The original TED Gestures dataset \cite{yoon2020speech, yoon2019robots} only contains 10 upper body joints without elaborate fingers of two hands. We newly collect a TED Hands dataset based on the raw videos of TED talking speeches. The videos are captured from the official TED channel on YouTube\footnote{We obey the TED Talks Team’s Creative Commons License (CC BY-NC-ND 4.0 International). In this work, all the videos from TED talking speeches are only used for research.}. 
To obtain reliable 3D hand joints and their corresponding upper body skeletons, we leverage a state-of-the-art 3D human pose estimator Fankmocap \cite{Rong_2021_ICCV} for annotation. In particular, we acquire 8 upper body joints and 30 figure joints in our dataset.

Concretely, we split the videos into 64-frame sequences under the following criteria:
\begin{itemize}
    \item  The above-mentioned 38 joints are visible for more than 48 frames in a sequence. Then, we interpolate the sequence to 64 frames.
    \vspace{-0.6em}
    \item  Since there might be multiple speakers in a single video of TED talking speeches (\emph{e.g.,} conversation between two speakers). To guarantee the continuity of body-hand movements, we only select the sequence that the joints of 64 frames belonging to a single speaker.
\end{itemize}

Finally, we obtain 1,755 videos with 134,456 sequences in our TED Hands dataset.
The statistics of B2H, TED Gesture, and our TED Hands datasets are reported in Tab. \ref{tab:statistics}. For our TED Hands dataset, the numbers of sequences in each data partition are: 
\begin{itemize}
    \item  Training set: 94,125.
    \vspace{-0.6em}
    \item  Validation set: 13,446.
    \vspace{-0.6em}
    \item  Testing set: 26,885.
\end{itemize}

\section*{D. Additional Visualization Results}
Here, we provide more visual results of our method as well as other competitors in the demo video. For more details, please refer to our \href{https://github.com/XingqunQi-lab/Diverse-3D-Hand-Gesture-Prediction}{\textit{project page}}.
Since all the comparison methods are designed without the diversification setting, we divide the comparisons into two parts. In the first part, we visualize the results of various competitors and the initial predictions of our method. In the second part, we visualize the diversification results based on our initial prediction from stage one.

\begin{figure*}[t]
\begin{center}
\includegraphics[width=1\linewidth]{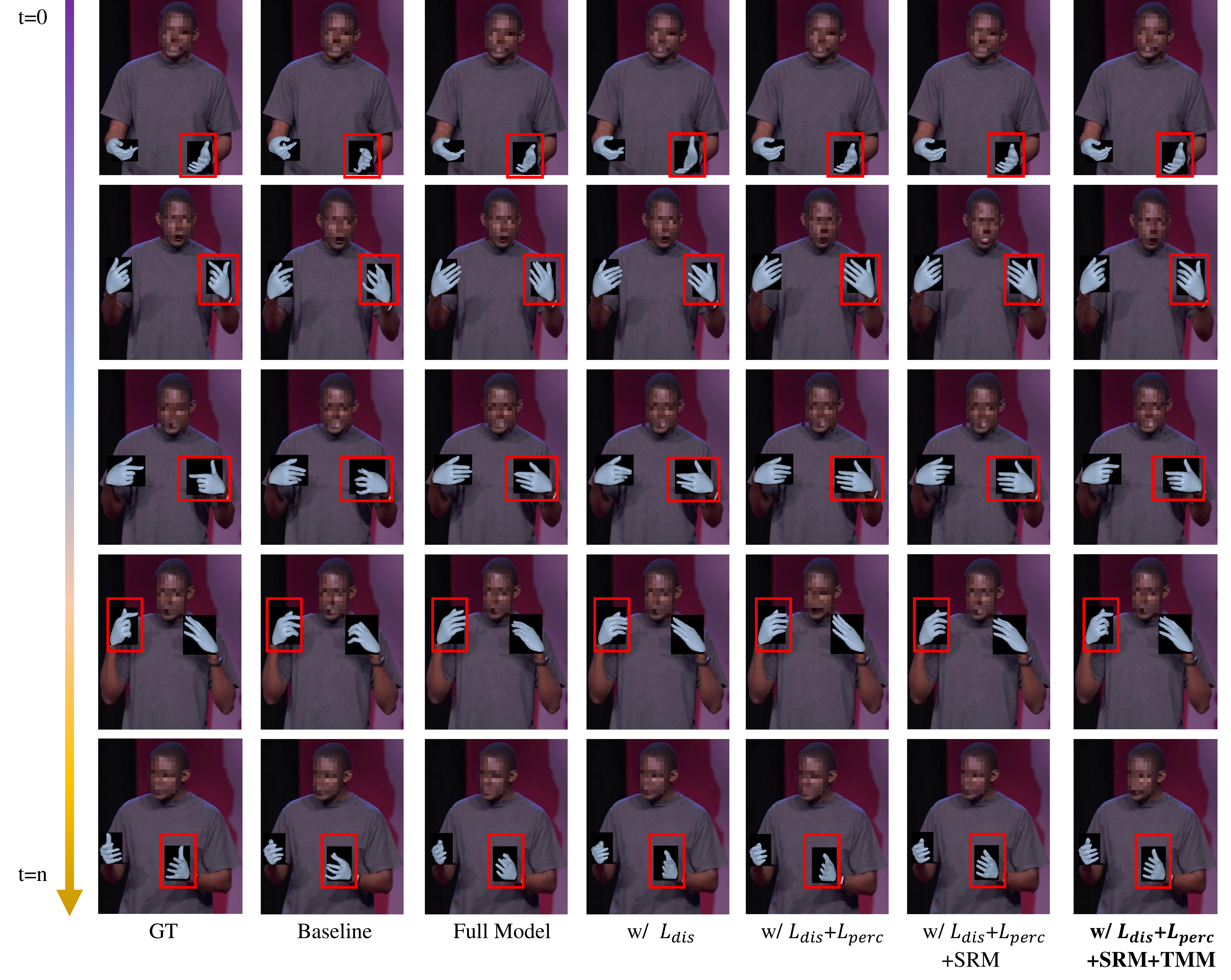}
\end{center}
\vspace{-1.5em}
\caption{Visual comparisons of ablation study on our newly collected TED Hands dataset. We show the key frames of the generated motions based on stage one initial predictions. Best view on screen.}
\label{fig:ablation2}
\end{figure*}

\begin{figure*}[t]
\begin{center}
\includegraphics[width=1\linewidth]{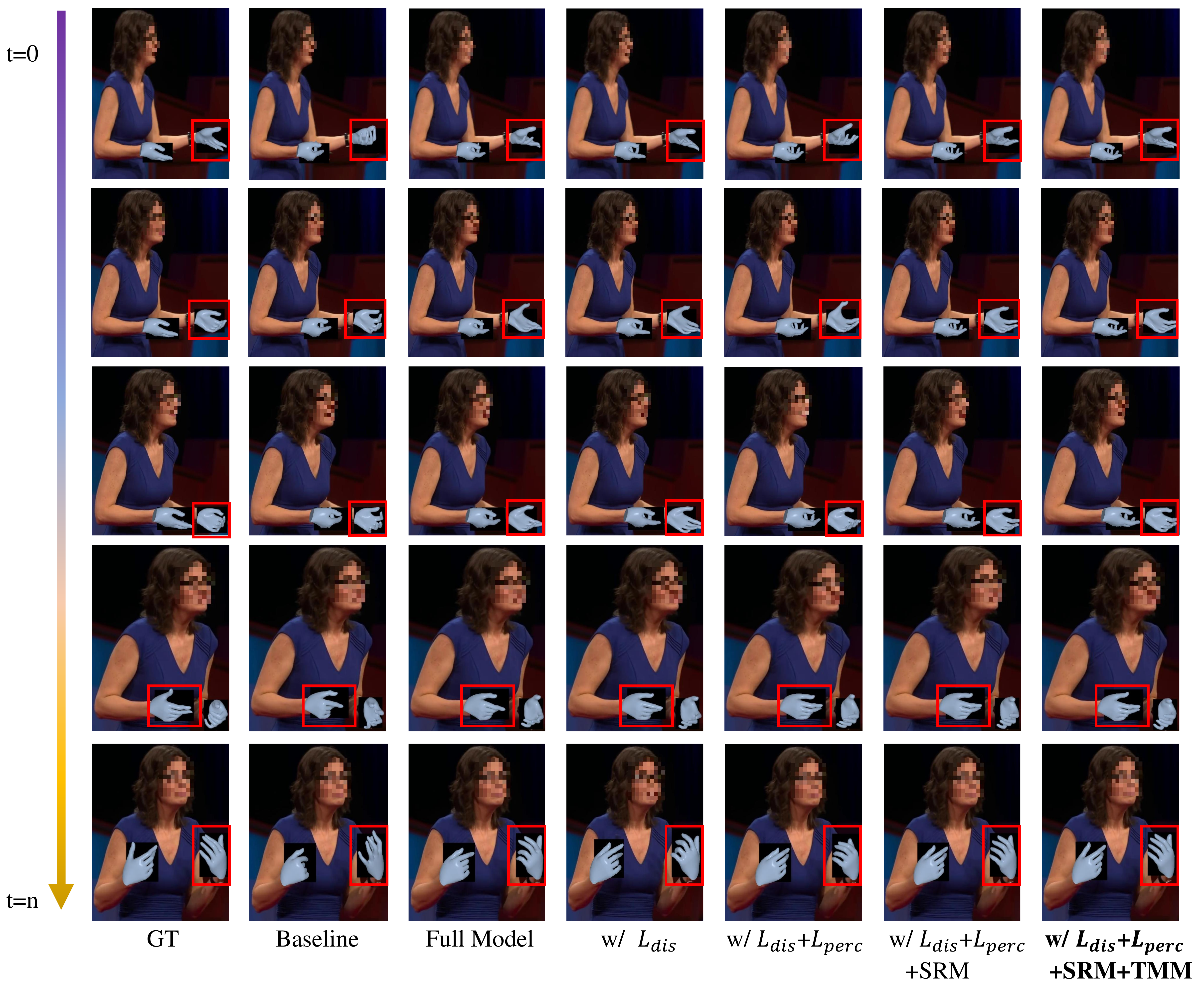}
\end{center}
\vspace{-1.5em}
\caption{Visual comparisons of ablation study on our newly collected TED Hands dataset. We show the key frames of the generated motions based on stage one initial predictions. Best view on screen.}
\label{fig:ablation1}
\end{figure*}

Moreover, to demonstrate the effectiveness of our proposed loss functions and components, we visualize vital frames of the generated motions based on stage one predictions. As illustrated in Fig. \ref{fig:ablation2}  and Fig. \ref{fig:ablation1}, we can clearly observe that all combinations of the different loss functions and components have positive impacts on 3D hand predictions.

\section*{E. Additional Results on Model Complexity}
We calculate the GFlops and inference time on a single NVIDIA RTX 2080 GPU, as reported in Tab.~\ref{tab:my-table2}. Due to the bilateral hand disentanglement process, the GFlops of our model are moderately higher than the second-best-performing method MRT. However, our method consistently outperforms other methods by large margins on L2, FHD, and MPJRE. 
The inference time of our model is around 32.921 ms (\emph{i.e.}, faster than 30 FPS). This inference speed allows our method to be deployed in real-time applications.

\vspace{-0.3em}
\begin{table}[t]
\centering
\caption{Comparison of model complexity, inference time, and performance on the TED Hands dataset. 
}
\vspace{-0.5em}
\label{tab:my-table2}
\footnotesize
\setlength{\tabcolsep}{1.6mm}{%
\renewcommand{\arraystretch}{0.98} 
\begin{tabular}{cccccc}
\toprule
Methods & GFlops $\downarrow$ & Time (ms) $\downarrow$ & L2 $\downarrow$ & FHD $\downarrow$ & MPJRE $\downarrow$ \\ \midrule
Body2hands & 0.068 & 2.823 & 2.551 & 1.174 & 11.371 \\
MRT & 0.211 & 4.341 & 2.325 & 0.877 & 10.314 \\
BTM & 0.052 & 11.089 & 2.350 & 1.111 & 10.440 \\
LTD & 0.113 & 5.962 & 2.482 & 1.367 & 11.078 \\
MotionMixer & 0.110 & 19.071 & 2.324 & 0.910 & 10.427 \\
SPGSN & 0.174 & 52.436 & 2.435 & 0.990 & 10.887 \\ \midrule
\rowcolor[HTML]{EFEFEF}
Ours & 0.503 & 32.921 & \textbf{2.037} & \textbf{0.258} & \textbf{8.888} \\ \bottomrule
\end{tabular}%
}
\end{table}

\end{document}